\documentclass[10pt,twocolumn,letterpaper]{article}

\usepackage[pagenumbers]{main}
\pdfoutput=1
\usepackage[table]{xcolor}
\usepackage{microtype}
\usepackage{graphicx}
\usepackage{booktabs, arydshln}
\usepackage{amsmath}
\usepackage{amssymb}
\usepackage{mathtools}
\usepackage{amsthm}

\usepackage{color}
\usepackage{enumitem}
\usepackage{fontawesome5}
\usepackage{multirow}
\usepackage{makecell}

\usepackage{algorithmic}
\usepackage{algorithm}
\usepackage{etoolbox,siunitx}

\definecolor{lblue}{RGB}{231, 66, 52}

\usepackage[pagebackref,breaklinks,colorlinks,citecolor=lblue]{hyperref}

\def\paperID{1}
\def\confName{RoboSense}
\def\confYear{IROS 2025}

\title{Towards Cross-Platform Generalization: Domain Adaptive 3D Detection with Augmentation and Pseudo-Labeling}

\author{%
Xiyan Feng$^{1}$, \quad Wenbo Zhang$^{1}$, \quad Lu Zhang$^{1}$\thanks{Corresponding author}, \quad Yunzhi Zhuge$^{1}$, \quad Huchuan Lu$^{1}$, \quad You He$^{2}$\\[0.3em]
{\small $^{1}$Dalian University of Technology}\\[-0.2em]
{\small $^{2}$Shenzhen International Graduate School, Tsinghua University}\\[-0.2em]
{\small\tt \{fxy,zwbo\}@mail.dlut.edu.cn, \{zhangluu,zgyz,lhchuan\}@dlut.edu.cn, heyou@tsinghua.edu.cn}
}

\begin{document}

\maketitle

\begin{abstract}
This technical report represents the award-winning solution to the Cross-platform 3D Object Detection task in the RoboSense2025 Challenge. Our approach is built upon PVRCNN++, an efficient 3D object detection framework that effectively integrates point-based and voxel-based features. On top of this foundation, we improve cross-platform generalization by narrowing domain gaps through tailored data augmentation and a self-training strategy with pseudo-labels. These enhancements enabled our approach to secure the 3rd place in the challenge, achieving a 3D AP of 62.67\% for the Car category on the phase-1 target domain, and 58.76\% and 49.81\% for Car and Pedestrian categories respectively on the phase-2 target domain.
\end{abstract}

\section{Introduction}
\label{sec:intro}

With the rise of autonomous driving technology, LiDAR-based 3D object detection has become a central research topic, providing essential environmental perception capabilities for tasks such as path planning and obstacle avoidance \cite{mao20223d,fernandes2021point,Zamanakos_Tsochatzidis_Amanatiadis_Pratikakis_2021,Arnold_Al-Jarrah_Dianati_Fallah_Oxtoby_Mouzakitis_2019,Qian_Lai_Li_2022}. By leveraging diverse feature representations including point-based\cite{Shi_Wang_Li_2019,Yang_Sun_Liu_Jia_2020,Pan_Xia_Song_Li_Huang_2021}, voxel-based\cite{Deng_Shi_Li_Zhou_Zhang_Li_2022,Yan_Mao_Li_2018}, and point-voxel based approaches\cite{Shi_Guo_Jiang_Wang_Shi_Wang_Li_2020,Shi_Jiang_Deng_Wang_Guo_Shi_Wang_Li_2023,Li_Wang_Wang_2021}, existing methods can effectively extract and encode geometric features from LiDAR point clouds, demonstrating remarkable accuracy on standard public datasets.

Building on these advances, the application scope of 3D object detection technology has expanded beyond traditional vehicles to diverse platforms such as drones and quadruped robots. Nevertheless, when detectors trained exclusively on vehicle platform datasets are directly deployed on these emerging platforms, they face a critical cross-platform generalization challenge: substantial performance degradation occurs due to significant domain shifts across platforms\cite{kong2025eventflyeventcameraperception,Zhou_2022}. Differences in motion patterns and mounting positions between ground vehicles, drones, and quadruped robots lead to distinct viewpoint variations, which in turn cause domain shifts in point density and spatial distribution. As a result, detectors trained on source domain data struggle to maintain robust performance in target domains, compromising the reliability of cross-platform deployment in real-world applications.

To address this challenge, we design a novel detection framework that extends the advanced PVRCNN++ detector\cite{Shi_Jiang_Deng_Wang_Guo_Shi_Wang_Li_2023} with specialized domain adaptation techniques. First, our approach implements the Cross-platform Jitter Alignment (CJA) augmentation method to mitigate point cloud distribution shifts caused by motion pattern variations across different platforms. By simulating viewpoint perturbations during training, CJA helps align geometric distributions between source and target domains. In addition, we incorporate the ST3D self-training paradigm\cite{Yang_Shi_Wang_Li_Qi_2021} to generate high-quality pseudo-labels on unlabeled target domain data, enabling the pre-trained 3D detector to progressively adapt to target domain through iterative refinement. This integrated framework preserves the strong detection capabilities of PVRCNN++ while substantially boosting its generalization ability across mobile platforms, offering a practical solution for efficient cross-platform application of models.

\begin{figure*}
  \centering
  \includegraphics[width=.9\linewidth]{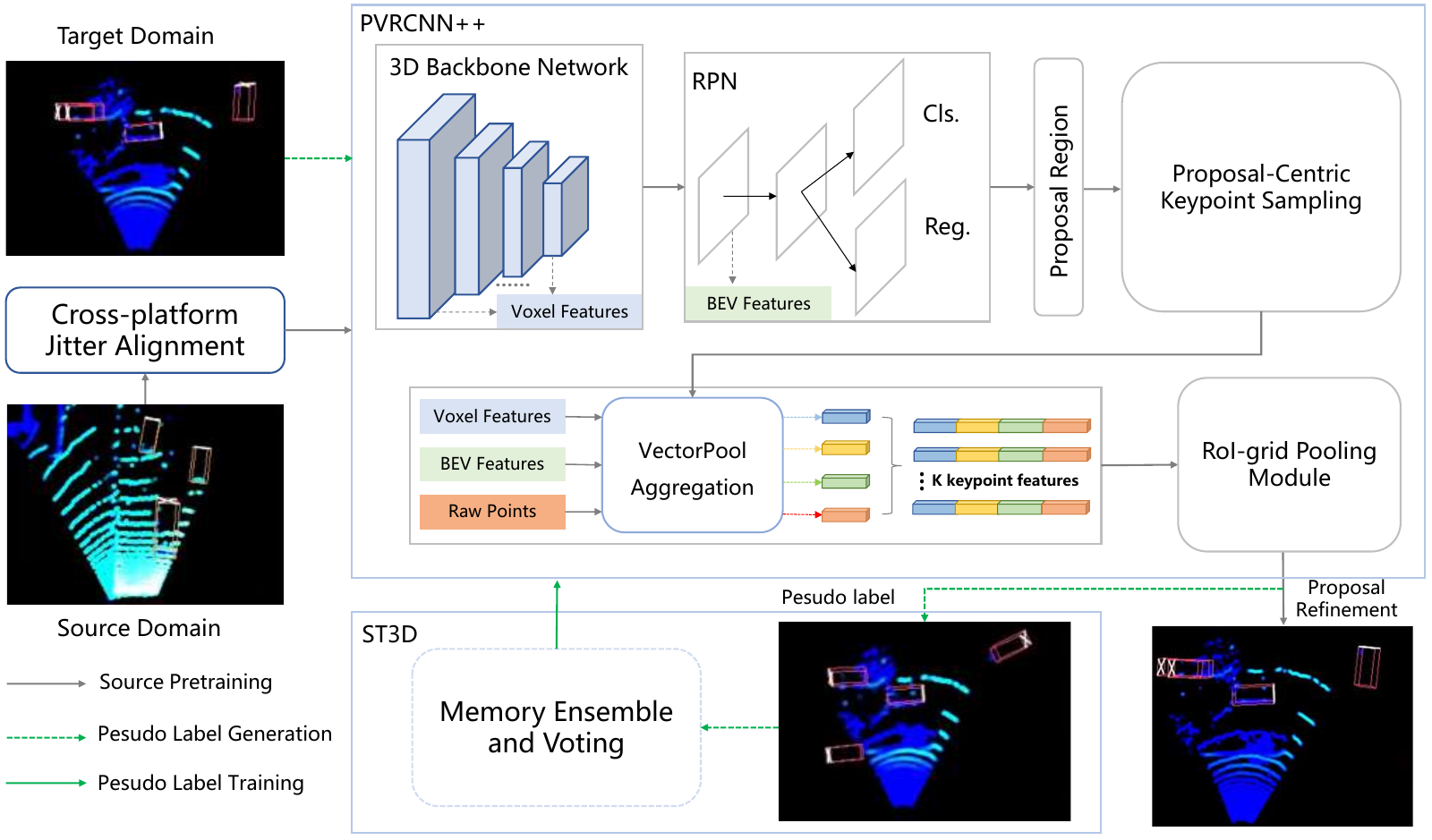}
  \caption{Overall pipeline of the proposed Cross-Platform 3D Detector. Source domain point clouds are processed by PVRCNN++ to produce detection results with labels, while target domain data generates pseudo-labels through the same detector. The ST3D module refines these pseudo-labels, and the CJA module operates exclusively on the source domain to enhance robustness by simulating platform-specific viewpoint jitter. Both source labels and refined target pseudo-labels are then utilized to iteratively update the model, enabling progressive adaptation across platforms without additional annotations.}
  \label{fig:single}
\end{figure*}
\section{Method}

\subsection{Cross-domain 3D object detection paradigm}

\textbf{3D Detector.}
We adopt an advanced two-stage 3D object detection framework based on PVRCNN++, which seamlessly integrates the advantages of both point-based and voxel-based detection paradigms. This framework achieves efficient feature extraction from 3D scenes through voxelization while preserving the fine-grained geometric information inherent in raw point clouds, thereby striking a balance between accuracy and efficiency.

In the first stage, PVRCNN++ transforms irregular point cloud data into structured voxel grids via Voxel Feature Encoding. A 3D sparse convolutional network then extracts multi-scale voxel features, which are used to generate high-quality proposals with preliminary classification and bounding box regression. We observed that the CenterHead, used for proposal generation, relies on heatmap peaks from the BEV feature map to localize bounding box centers. However, under large viewing angles, the point cloud density of small objects deteriorates significantly, causing heatmap peaks to frequently localize to ground surfaces or empty voxels. This phenomenon leads to center drift and degraded recall rates. To combat this issue, we replace the CenterHead with an AnchorHead that incorporates predefined anchors of diverse scales and orientations. This modification provides enhanced geometric priors to the model, significantly improving proposal quality while ensuring greater stability and efficiency during training. 

In the second stage, representative keypoints are first extracted from the global scene, and the Voxel Set Abstraction (VSA) module aggregates rich contextual voxel features onto these keypoints. RoI-grid pooling is subsequently applied to perform structured grid point sampling within each 3D proposal region, where high-quality VSA features are aggregated. Finally, a proposal refinement network processes these features to produce refined bounding box predictions.

\noindent\textbf{Unsupervised Domain Adaptation.}
Cross-platform point clouds exhibit significant geometric distribution disparities due to viewpoint variations. For instance, different mobile platforms may introduce distinct LiDAR vibration patterns during carrier motion, and substantial variations in platform elevation lead to completely different absolute coordinates for identical objects across platforms. Consequently, models trained solely on the source domain often suffer severe performance degradation or even failure when directly applied to target domains.

To address this challenge, we employ ST3D, an unsupervised domain adaptation method built on a self-training strategy. ST3D iteratively generates pseudo-labels for unlabeled target domain data and uses them to fine-tune the model pre-trained on the source domain. This iterative process encourages the model to learn domain-invariant feature representations, thereby enhancing robustness across heterogeneous sensor setups and environments.

\subsection{Data Augmentation}
In cross-platform 3D detection tasks, different mobile platforms induce distinct jitter characteristics in their mounted LiDAR sensors due to variations in motion patterns and mechanical structures. The source domain vehicle platform typically operates on flat road surfaces, where changes in pitch and roll angles are relatively small. As a result, the model trained on such data learns feature representations constrained to a narrow range of viewpoints and motion priors, limiting its ability to generalize to novel motion patterns of target platforms. 

Inspired by\cite{liang2025perspectiveinvariant3dobjectdetection}, we implement the Cross-platform Jitter Alignment (CJA) augmentation technique to explicitly compensate for jitter distribution discrepancies across platforms. By introducing controlled pose perturbations during source domain pre-training, CJA encourages the detector to learn feature representations invariant to platform-specific jitter, thereby enhancing the model's cross-platform generalization capability.
Concretely, for each training sample, we uniformly sample pitch increments $\Delta\theta$ and roll increments $\Delta\phi$ from a predefined jitter angle range. A composite rotation matrix $\mathbf{R}(\Delta\phi, \Delta\theta)$ is then applied to the entire point cloud scene.

To preserve annotation consistency, the centers of all bounding boxes undergo the same transformation. The dimensions and orientation angles of the boxes remain fixed, while only their spatial positions are adjusted to align with the rotated point cloud. This process maintains geometric consistency of the annotations while effectively simulating viewpoint changes induced by platform motion in real-world LiDAR data.

\subsection{Training framework}
We adopt a systematic two-stage training framework to address domain adaptation in cross-platform 3D detection. This framework first establishes a strong foundational detection capability on the source domain, followed by fine-tuning on the target domain using a self-training strategy. This phased design ensures good initial performance and facilitates effective adaptation to the data distribution of the target domain.

In stage 1, the objective is to build a robust baseline detector that provides high-quality initial weights for subsequent domain adaptation. To achieve this objective, we train the PVRCNN++ detector on the source domain data using a standard supervised learning paradigm, which facilitates the learning of discriminative features from annotated LiDAR point clouds.


In stage 2, building upon the pre-trained weights from the first stage, we employ the ST3D self-training method for domain adaptation fine-tuning on unlabeled target domain data. Through an iterative pseudo-label optimization process, the model gradually reduces the inter-domain discrepancy, enabling it to adapt to the data distribution characteristics of the target domain.We adopt differentiated threshold design to ensure that high-quality pseudo-labels can be generated in all categories. 


\begin{table*}
  \centering
  \begin{tabular}{@{}cccccccc@{}}
    \toprule
    Detector & CJA & ST3D & \textbf{Car AP@0.5} & Car AP@0.7 & \textbf{Ped. AP@0.5} & Ped. AP@0.25 & Score \\
    \midrule
    \multirow{3}{*}{PointRCNN\cite{Shi_Wang_Li_2019}} & - & - & 46.29 & 25.71 & 41.17 & 44.74 & 43.73\\
    & \checkmark & - & 46.04 & 26.83 & 40.97 & 44.90 & 43.50\\
    & \checkmark & \checkmark & 46.46 & 26.45 & 27.92 & 31.34 & 37.19\\
    \midrule
    \multirow{3}{*}{VoxelRCNN\cite{Deng_Shi_Li_Zhou_Zhang_Li_2022}} & - & - & 26.95 & 10.88 & 28.44 & 33.24 & 27.70\\
    & \checkmark & - & 40.52 & 22.38 & 45.18 & 49.34 & 42.85\\
    & \checkmark & \checkmark & 45.43 & 26.09 & 48.03 & 51.95 & 46.73\\
    \midrule
    \multirow{3}{*}{PDV\cite{hu2022pointdensityawarevoxelslidar}} & - & - & 26.12 & 11.17 & 26.27 & 30.27 & 26.20\\
    & \checkmark & - & 43.39 & 24.31 & 46.11 & 50.84 & 44.75\\
    & \checkmark & \checkmark & 44.53 & 25.23 & 47.01 & 51.86 & 46.28\\
    \midrule
    \multirow{3}{*}{PVRCNN++\cite{Shi_Jiang_Deng_Wang_Guo_Shi_Wang_Li_2023}} & - & - & 29.44 & 9.98 & 14.94 & 18.16 & 22.19 \\
    & \checkmark & - & 43.94 & 23.96 & 46.83 & 52.28 & 45.39\\
    & \checkmark & \checkmark & 54.72 & 29.41 & 48.25 & 54.76 & 51.48 \\
    \hdashline
    PVRCNN++$^{*}$ & \checkmark & \checkmark & \textbf{58.79} & \textbf{30.89} & \textbf{49.81} & \textbf{55.27} & \textbf{54.29} \\
    \bottomrule
  \end{tabular}
  \caption{Performance comparison of different 3D detection frameworks under cross-platform adaptation from vehicle to quadruped robot platforms. Symbol * denotes replacing the RPNHead with AnchorHead. All scores are given in percentage (\%). We report the 3D Average Precision (AP) for Cars at IoU thresholds of 0.5 and 0.7, and for Pedestrians at thresholds of 0.25 and 0.5.}
  \label{tab:1}
\end{table*}

\section{Experiments}

\subsection{Experimental Setup}

Our experiments were conducted on the Track5 dataset released in RoboSense2025, following the official challenge protocol for data preparation. The dataset consists of three subsets: source domain data, Phase 1 target domain data, and Phase 2 target domain data. These subsets contain point cloud-image pairs collected by LiDAR and camera systems mounted on vehicles, drones, and quadrupeds, respectively. Point cloud annotations are available exclusively for the source domain. The training consists of two stages: model pre-training is performed solely on the source domain data; the subsequent self-training stage generates pseudo-labels on the target domain data and uses them for training, with evaluation conducted on the target domain sets.

\subsection{Implementation Details}

Our model is implemented based on the OpenPCDet codebase\cite{openpcdet2020} using PyTorch. Both the pre-training and self-training stages are conducted on 4 NVIDIA GTX 4090 GPUs, with a batch size of 4 per GPU. We adopt the AdamW optimizer with OneCycle learning rate policy. The initial learning rate is set to 0.01 for pre-training and $1.5\times10^{-3}$ for self-training
. The self-training stage runs for 5 epochs, with pseudo-labels updated every 4 epochs. For Phase 1 data, the confidence threshold is set to 0.7 and the negative sample threshold to 0.2. For Phase 2 data, the confidence thresholds are 0.85 for the Car class and 0.55 for the Pedestrian class, with a uniform negative sample threshold of 0.20.

\subsection{Ablation Study}

We conduct comprehensive ablation studies to evaluate the effectiveness of our proposed model and its compatibility with different detector architectures. As shown in Table~\ref{tab:1}, the baseline PVRCNN++ achieves 29.44\% Car AP@0.5 and 14.94\% Pedestrian AP@0.5. Incorporating CJA augmentation improves these metrics to 43.94\% and 46.83\% respectively, demonstrating its effectiveness in enhancing the model's robustness against platform-specific jitter through geometric alignment. Further combining with ST3D self-training elevates performance to 54.72\% Car AP@0.5 and 48.25\% Pedestrian AP@0.5, reflecting its capability to progressively adapt the model to target domain characteristics through iterative pseudo-label refinement. Finally, replacing the original RPN head with our AnchorHead modification establishes the optimal model, reaching 58.79\% Car AP@0.5 and 49.81\% Pedestrian AP@0.5, demonstrating the critical role of geometric priors in enhancing proposal quality for cross-platform detection.

Beyond evaluating our primary framework, we further analyze the generalizability of CJA and ST3D across different types of detectors. The results show that both components significantly benefit voxel-based and point-voxel hybrid methods while showing limited effectiveness on pure point-based architectures. Specifically, both VoxelRCNN and PDV demonstrate substantial performance gains through the application of CJA and ST3D modules, with approximately 20\% improvement in Car AP@0.5 and around 18\% gain in Pedestrian AP@0.5. In contrast, PointRCNN remains largely unaffected by CJA and even experiences performance degradation in pedestrian detection when applying ST3D. This architectural sensitivity indicates that our approach is particularly suitable for detectors utilizing voxel representations.



\section{Conclusion}

In order to enhance cross-platform detection accuracy, we improved our competition framework by implementing a cross-platform 3D detection system built upon the PVRCNN++ detector pre-trained on source domain data. This framework incorporates the CJA data augmentation technique to explicitly mitigate geometric distribution discrepancies across platforms, and is further enhanced by the ST3D self-training paradigm that generates high-quality pseudo-labels for effective domain adaptation. Experimental results demonstrate that our improvements achieve remarkable performance gains in cross-platform scenarios.

{
\small
\bibliographystyle{unsrt}
\bibliography{main}

\begin{thebibliography}{10}

\bibitem{mao20223d}
Jiageng Mao, Shaoshuai Shi, Xiaogang Wang, and Hongsheng Li.
\newblock 3d object detection for autonomous driving: A review and new outlooks.
\newblock {\em arXiv preprint arXiv:2206.09474}, 1(1):1, 2022.

\bibitem{fernandes2021point}
Duarte Fernandes, Ant{\'o}nio Silva, Rafael N{\'e}voa, Cl{\'a}udia Sim{\~o}es, Dibet Gonzalez, Miguel Guevara, Paulo Novais, Jo{\~a}o Monteiro, and Pedro Melo-Pinto.
\newblock Point-cloud based 3d object detection and classification methods for self-driving applications: A survey and taxonomy.
\newblock {\em Information Fusion}, 68:161--191, 2021.

\bibitem{Zamanakos_Tsochatzidis_Amanatiadis_Pratikakis_2021}
Georgios Zamanakos, Lazaros Tsochatzidis, Angelos Amanatiadis, and Ioannis Pratikakis.
\newblock A comprehensive survey of lidar-based 3d object detection methods with deep learning for autonomous driving.
\newblock {\em Computers \& Graphics}, page 153–181, Oct 2021.

\bibitem{Arnold_Al-Jarrah_Dianati_Fallah_Oxtoby_Mouzakitis_2019}
Eduardo Arnold, Omar~Y. Al-Jarrah, Mehrdad Dianati, Saber Fallah, David Oxtoby, and Alex Mouzakitis.
\newblock A survey on 3d object detection methods for autonomous driving applications.
\newblock {\em IEEE Transactions on Intelligent Transportation Systems}, page 3782–3795, Oct 2019.

\bibitem{Qian_Lai_Li_2022}
Rui Qian, Xin Lai, and Xirong Li.
\newblock 3d object detection for autonomous driving: A survey.
\newblock {\em Pattern Recognition}, page 108796, Oct 2022.

\bibitem{Shi_Wang_Li_2019}
Shaoshuai Shi, Xiaogang Wang, and Hongsheng Li.
\newblock Pointrcnn: 3d object proposal generation and detection from point cloud.
\newblock In {\em 2019 IEEE/CVF Conference on Computer Vision and Pattern Recognition (CVPR)}, Jun 2019.

\bibitem{Yang_Sun_Liu_Jia_2020}
Zetong Yang, Yanan Sun, Shu Liu, and Jiaya Jia.
\newblock 3dssd: Point-based 3d single stage object detector.
\newblock In {\em 2020 IEEE/CVF Conference on Computer Vision and Pattern Recognition (CVPR)}, Jun 2020.

\bibitem{Pan_Xia_Song_Li_Huang_2021}
Xuran Pan, Zhuofan Xia, Shiji Song, Li~Erran Li, and Gao Huang.
\newblock 3d object detection with pointformer.
\newblock In {\em 2021 IEEE/CVF Conference on Computer Vision and Pattern Recognition (CVPR)}, Jun 2021.

\bibitem{Deng_Shi_Li_Zhou_Zhang_Li_2022}
Jiajun Deng, Shaoshuai Shi, Peiwei Li, Wengang Zhou, Yanyong Zhang, and Houqiang Li.
\newblock Voxel r-cnn: Towards high performance voxel-based 3d object detection.
\newblock {\em Proceedings of the AAAI Conference on Artificial Intelligence}, page 1201–1209, Sep 2022.

\bibitem{Yan_Mao_Li_2018}
Yan Yan, Yuxing Mao, and Bo~Li.
\newblock Second: Sparsely embedded convolutional detection.
\newblock {\em Sensors}, page 3337, Oct 2018.

\bibitem{Shi_Guo_Jiang_Wang_Shi_Wang_Li_2020}
Shaoshuai Shi, Chaoxu Guo, Li~Jiang, Zhe Wang, Jianping Shi, Xiaogang Wang, and Hongsheng Li.
\newblock Pv-rcnn: Point-voxel feature set abstraction for 3d object detection.
\newblock In {\em 2020 IEEE/CVF Conference on Computer Vision and Pattern Recognition (CVPR)}, Jun 2020.

\bibitem{Shi_Jiang_Deng_Wang_Guo_Shi_Wang_Li_2023}
Shaoshuai Shi, Li~Jiang, Jiajun Deng, Zhe Wang, Chaoxu Guo, Jianping Shi, Xiaogang Wang, and Hongsheng Li.
\newblock Pv-rcnn++: Point-voxel feature set abstraction with local vector representation for 3d object detection.
\newblock {\em International Journal of Computer Vision}, page 531–551, Feb 2023.

\bibitem{Li_Wang_Wang_2021}
Zhichao Li, Feng Wang, and Naiyan Wang.
\newblock Lidar r-cnn: An efficient and universal 3d object detector.
\newblock In {\em 2021 IEEE/CVF Conference on Computer Vision and Pattern Recognition (CVPR)}, Jun 2021.

\bibitem{kong2025eventflyeventcameraperception}
Lingdong Kong, Dongyue Lu, Xiang Xu, Lai~Xing Ng, Wei~Tsang Ooi, and Benoit~R. Cottereau.
\newblock Eventfly: Event camera perception from ground to the sky, 2025.

\bibitem{Zhou_2022}
Kaiyang Zhou, Ziwei Liu, Yu~Qiao, Tao Xiang, and Chen~Change Loy.
\newblock Domain generalization: A survey.
\newblock {\em IEEE Transactions on Pattern Analysis and Machine Intelligence}, page 1–20, 2022.

\bibitem{Yang_Shi_Wang_Li_Qi_2021}
Jihan Yang, Shaoshuai Shi, Zhe Wang, Hongsheng Li, and Xiaojuan Qi.
\newblock St3d: Self-training for unsupervised domain adaptation on 3d object detection.
\newblock In {\em 2021 IEEE/CVF Conference on Computer Vision and Pattern Recognition (CVPR)}, Jun 2021.

\bibitem{liang2025perspectiveinvariant3dobjectdetection}
Ao~Liang, Lingdong Kong, Dongyue Lu, Youquan Liu, Jian Fang, Huaici Zhao, and Wei~Tsang Ooi.
\newblock Perspective-invariant 3d object detection, 2025.

\bibitem{hu2022pointdensityawarevoxelslidar}
Jordan S.~K. Hu, Tianshu Kuai, and Steven~L. Waslander.
\newblock Point density-aware voxels for lidar 3d object detection, 2022.

\bibitem{openpcdet2020}
OpenPCDet~Development Team.
\newblock Openpcdet: An open-source toolbox for 3d object detection from point clouds.
\newblock \url{https://github.com/open-mmlab/OpenPCDet}, 2020.

\end{thebibliography}
}

\end{document}